\pdfoutput=1

\documentclass[11pt]{article}

\usepackage{EACL2023}

\usepackage{times}
\usepackage{latexsym}
\usepackage{amsmath}

\usepackage[T1]{fontenc}

\usepackage{url}

\usepackage{devanagari}

\usepackage{listings}
\usepackage{microtype}
\usepackage{booktabs}
\usepackage{fontawesome}
\usepackage{wrapfig}
\usepackage{graphicx}
\graphicspath{ {./figures/} }

\newcommand{\sm}[1]{\textcolor{red}{#1}} %
\newcommand{\sai}[1]{\textcolor{green!80!black}{#1}} %

\newcommand{\xlmr}{XLM-R$_{\textrm{base}}$}
\newcommand{\ma}{MultiATIS++{}}
\newcommand{\ms}{MultiSNIPS{}}
\newcommand{\wa}{WikiANN{}}
\newcommand{\nli}{XNLI{}}

\title{Robustification of Multilingual Language Models to Real-world Noise in Crosslingual Zero-shot Settings with Robust Contrastive Pretraining}

\usepackage{cleveref}
\crefformat{section}{\S#2#1#3} %
\crefformat{subsection}{\S#2#1#3}
\crefformat{subsubsection}{\S#2#1#3}

\author{Asa Cooper Stickland$^{\star\mathparagraph\dagger}$, Sailik Sengupta$^{\star\ddagger}$, Jason Krone$^{\mathparagraph\ddagger}$, He He$^{\ddagger \diamondsuit}$, Saab Mansour$^\ddagger$ \\[4pt]
  $^\dagger$University of Edinburgh, $^\ddagger$\faAmazon WS AI Labs, $^\diamondsuit$New York University\\[1pt]
  {\small \texttt{a.cooper.stickland@ed.ac.uk,\{sailiks,saabm,hehea\}@amazon.com}}
}

\begin{document}

\maketitle

\begin{abstract}

Advances in neural modeling have achieved state-of-the-art (SOTA) results on public natural language processing (NLP) benchmarks, at times surpassing human performance. However, there is a gap between public benchmarks and real-world applications where {\em noise}, such as typographical or grammatical mistakes, is abundant and can result in degraded performance. Unfortunately, works which evaluate the robustness of neural models on noisy data and propose improvements, are limited to the English language. Upon analyzing noise in different languages, we observe that noise types vary greatly across languages. Thus, existing investigations do not generalize trivially to multilingual settings. To benchmark the performance of pretrained multilingual language models, we construct noisy datasets covering five languages and four NLP tasks and observe a clear gap in the performance between clean and noisy data in the zero-shot cross-lingual setting. After investigating several ways to boost the robustness of multilingual models in this setting, we propose Robust Contrastive Pretraining (RCP). RCP combines data augmentation with a contrastive loss term at the pretraining stage and achieves large improvements on noisy (\& original test data) across two sentence-level (+3.2\%) and two sequence-labeling (+10 F1-score) multilingual classification tasks.
\let\thefootnote\relax\footnote{$^\star$ Equal Contribution, $^\mathparagraph$ Work done while at Amazon.}
\end{abstract}

\section{Introduction}

Recently, multilingual pre-trained language models like mBERT \cite{devlin-etal-2019-bert}, XLM-R \cite{conneau2020unsupervised} and various others \cite{chi-etal-2021-infoxlm, xue-etal-2021-mt5, chi-etal-2022-xlm} have improved multilingual language understanding by pretraining large Transformer models on web-scale corpora (such as Wikipedia, Common-Crawl). These models achieve state-of-the-art performance on cross-lingual transfer and many multilingual NLP tasks \cite{wu-dredze-2019-beto, pires-etal-2019-multilingual}. However, a real-world system will encounter {\em real-world noise}, such as linguistic variations and common errors observed in textual data, that are often absent from benchmark datasets.

While prior works focused on this issue of robustness in monolingual settings \cite{peng-etal-2021-raddle,sengupta-etal-2021-robustness,tan-etal-2020-morphin}, investigation has been scarce for multilingual settings. In this paper, we study the effect of realistic noise in multilingual settings and propose methods to boost the robustness of multilingual language models across four NLP tasks: Intent Classification (IC), Slot Labeling (SL), Named Entity Recognition (NER) and Natural Language Inference (NLI).

\begin{figure*}
    \centering
    \includegraphics[width=0.99\textwidth]{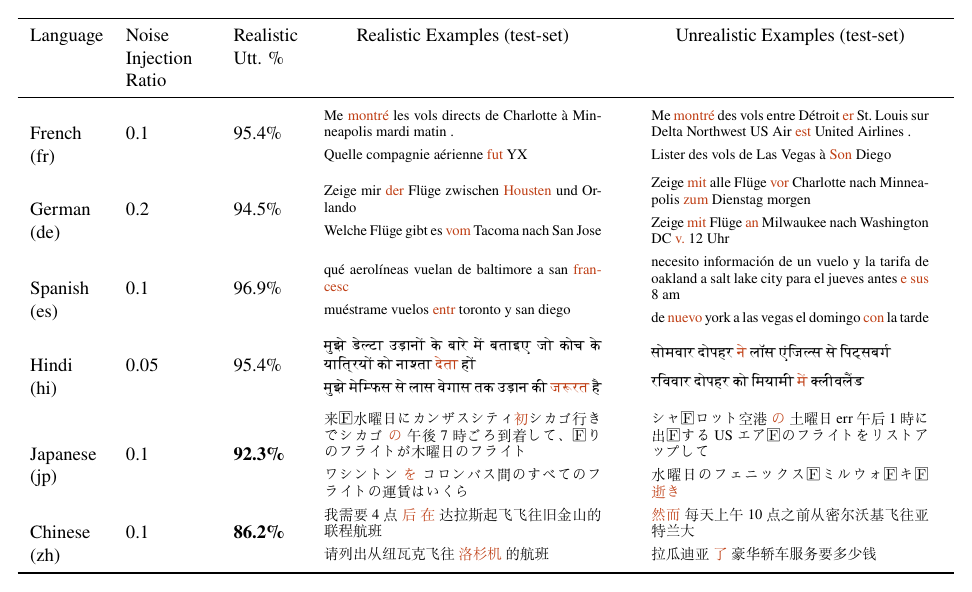}
    \caption{MultiATIS++ test set injected with real-world noise mined from Wikipedia edits. The highest error injection ratio found to be realistic by human experts is shown alongside the realistic utterance percentage. We do not include the noisy test sets for Chinese and Japanese in our analysis owing to low ($<95\%$) realism.}
    \label{tab:noise_data_analysis}
    \vspace{-5pt}
    \label{fig:my_label}
\end{figure*}

Due to the lack of multilingual noisy \textit{evaluation} data, we synthesize benchmarks by mining noise from publicly available corpora and injecting them into the test sets associated with each of the four tasks. We conduct human validation to ensure that this noised data is indeed realistic (see examples from \ma~in \autoref{tab:noise_data_analysis}) and identify the variety of noise-types seen across languages (in \cref{sec:ds_creation}). These analyses highlight the potential of our test-set in evaluating (and motivating future research on) multilingual robustness.

To benchmark the performance of multilingual systems, we consider accuracy metrics on two utterance-level tasks (IC\% and NLI\%) and F1-scores on two token-level classification tasks (SL-F1 and NER-F1). Specifically, we seek to evaluate the model’s performance on the noised version of the test datasets in a {\em zero-shot cross-lingual setting}. In this scenario, we have training data for a task available only in one language (in our case, English) and test-data in various languages \cite{liu-etal-2019-zero,liu2020attention}.

While training data augmentation increases model robustness for monolingual (i.e. English) settings, it is not immediately obvious if these robustness gains can transfer across languages, as error types can often be language-specific. For example, typos in Devanagari script can differ from those seen in Latin scripts (e.g. {\dn -\8{k}l} $\rightarrow$ {\dn s{k}ul} in Devanagari showcases that a joined character is incorrectly separated into two characters in the word `school’).

Thus, to improve the robustness of pretrained multilingual models across noise in all languages, we propose Robust Constrastive Pretraining (RCP) that couples multilingual noisy data-augmentation with a contrastive learning loss term during pretraining; this encourages the model to develop similar representations for the original and the noised version of a sentence.

On the noisy test sets, our method improves the multilingual model performance across all metrics and multilingual tasks-- IC\% by $4.9\%$ on \ma, $4.1\%$ on \ms; SL-F1 by $18.4$ on \ma, $8.6$ on \ms; NER-F1 by $2.9$ on \wa; NLI\% by $0.7\%$ on \nli.
In summary, our primary contributions are:

\begin{enumerate}

    \item We construct multilingual test data to evaluate the robustness of NLP models to noise (\cref{sec:ds_creation}).

    \item We show that the performance of existing multilingual language models deteriorates on four tasks when tested on the noisy test data  (\cref{sec:robustness-models}).

    \item We introduce Robust Contrastive Pretraining (RCP) to boost the robustness of existing multilingual language models (\cref{sec:rob_con}).

\end{enumerate}%

Our code and data is available on \href{https://github.com/amazon-science/multilingual-robust-contrastive-pretraining}{\textit{Github (repo: amazon-science/multilingual-robust-contrastive-pretraining)}} .

\section{Related Work}
Many prior works demonstrate the brittleness of neural models on different noise phenomena such as misspellings \cite{belinkov2017synthetic,karpukhin2019training,moradi-etal-2021-measuring}, casing variation \cite{van-miltenburg-etal-2020-evaluation}, paraphrases \cite{einolghozati2019improving}, morphological variance \cite{tan-etal-2020-morphin}, synonyms \cite{sengupta-etal-2021-robustness}, and dialectical variance \cite{Sarkar2022}. A popular approach to improve the robustness to noise is fine-tuning models with data augmentation \cite{feng-etal-2021-survey} at either the pretraining \cite{tan-etal-2020-morphin,Sarkar2022} or the task-training stage \cite{peng-etal-2021-raddle}. 
These works consider monolingual pre-trained models and primarily focus on English. While recent works on token-free models motivate robustness in multilingual settings \cite{DBLP:journals/corr/abs-2103-06874,xue-etal-2022-byt5,tay2021charformer}, {\em examining the robustness of SOTA multilingual pre-trained models (and improving them) remains unexplored}. Hence, we investigate-- (1) are multilingual models robust to noise seen in different languages (that may be dissimilar to noise types seen in English)? (2) can we get and leverage multi-lingual noise data to improve multilingual models? and (3) do automatic data-augmentation methods designed for English improve robustness to multilingual noise?

To boost the robustness of multilingual models to diverse multilingual noise, we leverage multilingual data augmentation at the pretraining stage and use contrastive learning. Our effort complements work in computer vision that showcases contrastive learning with adversarial learning at task-training \cite{fan2021does,ghosh2021contrastive} and pre-training time \cite{jiang2020robust,kim2020adversarial} can improve model robustness. NLP has also seen a plethora of work that leverages contrastive learning, but seldom to alleviate robustness concerns \cite{jaiswal2020survey}. Similar concepts, such as Adversarial Logit Pairing \cite{einolghozati2019improving}, used at task-training time have proven to be less effective than data augmentation approaches \cite{sengupta-etal-2021-robustness} in boosting robustness.

All the aforementioned works lack in at least one of the two novel aspects of this paper-- robustness to real-world (as opposed to adversarial) noise, and/or multilinguality.
Lastly, the aspect of cross-lingual knowledge transfer has been studied in the context of different NLP tasks; typically, from a high-resource language to a low-resource one, as exemplified by the XTREME benchmark \cite{hu2020xtreme}. In this paper, we investigate the cross-lingual transferability of robustness to real-world noise.

\section{Constructing Noisy Test Data}

\label{sec:ds_creation}

As no existing benchmarks exist to evaluate the robustness of multilingual models, we construct noisy test sets in multiple languages for four tasks. First, we construct a word-level error-and–correction dictionary by leveraging the Wikipedia edit corpora. Then, we sample replacements from this dictionary and inject them into the test data for the various multilingual tasks, focusing on replacements that only affect individual words but do not change word order. Finally, we conduct human evaluation to filter out test sets that are not deemed to be realistic by language experts.

\subsection{Wiki-edit Mining}
\label{sec:wiki_mining}

Wikipedia\footnote{\url{https://meta.wikimedia.org/wiki/List\_of\_Wikipedias}} is a public encyclopedia available in multiple languages. Wikipedia editors create and iteratively edit its contents. We leverage these edits to construct error-correction word dictionaries (later used to create noisy test data). Our approach to mining edits is similar to \citet{tanaka-etal-2020-building}, but we consider multiple languages (as opposed to only Japanese), and additionally create dictionaries of word-level edits. 

To isolate likely useful edits, we first consider each revision page of an article and split it into a list of sentences using NLTK \cite{bird2009natural}. Second, we filter out sentence pairs from two consecutive edit versions ensuring both sentences have (1) 2-120 tokens, (2) a difference if $<5$ tokens, and (3) a relative edit-distance within $30\%$ of the shorter sentence. Third, we leverage language-specific tokenizes {\tt difflib}\footnote{\url{https://docs.python.org/3/library/difflib.html}} to extract exact token-level deltas between the sentence pair. At last, we ensure word pairs (in these deltas) that have at least one character-level Levenshtein edit-distance from each other\footnote{For Chinese characters, including Kanji, even a single character distance could imply a different word.} and none of words are only numbers or punctuation tokens. Note that edits to Wikipedia involve changes to factual information, such as dates, rather than incorrect spelling or grammar; thus, the last step is necessary.

We can finally create a {\em noise dictionary} of correct-to-incorrect words that has frequency information about the different errors. For example, an element of the dictionary (in Spanish) looks like {\tt \{de: [(del, 0.52), (se, 0.32), (do, 0.1), (dë, 0.04), (en, 0.02)]\}}.

\subsection{Injecting Noise into Test sets}

We use the noise dictionaries to create a noised version of the original test data for the four tasks-- \ma~\cite{xu-etal-2020-end}, \ms, \wa~\cite{pan-etal-2017-cross} and \nli~\cite{conneau-etal-2018-xnli}. After tokenization, we sample tokens randomly without replacement. In each sampling step, we sample based on a uniform probability distribution over the individual tokens and then check if the token exists in the noise dictionary. If so, we replace it with a noised version from the dictionary; the noised version is sampled based on its probability in the noise dictionary (that is proportional to the frequency of its occurrence in the noisy corpora). This procedure continues till we noise a particular number of tokens, precisely between $1$ and $\min(4, pL)$  where $p$ a controllable fraction (chosen as a hyperparameter at first, and finalized based on human evaluation described in \cref{sec:human_verification}), and $L$ is the number of words in the sentence.

\subsection{Human Verification of Noised Test-sets}
\label{sec:human_verification}

During human evaluation, we analyse the noisy data created for the \ma~dataset. 
We asked the language expert to assume that a user who may not be a native speaker, or in a hurry, or sloppy, was trying to find out flight information via text chat, and evaluate realism with this in mind. Note that analysis of noise types for \ma~generalizes well to other datasets as we use the same error-correction dictionaries for injecting noise into all the test-sets.

Our language experts have graduate/doctoral degrees in linguistics, computational linguistics, or natural language processing and are fluent/native speakers of the respective languages. We employed the human experts and compensated them fairly to conduct this study (see \cref{sec:ethics} for details). The experts are given $45$ examples without being told that $15$ examples have $5\%$, $15$ have $10\%$, and $15$ have $20\%$ noised tokens and asked three questions about each example. (1) Is the noised sentence realistic, moderately realistic, or unrealistic? (2) What type of noise is present in the sentence (we supply an initial list and let them add more)? and (3) Are the intent and slot labels unchanged? Based on their initial feedback, we choose the most realistic noise fraction (i.e. $5$, $10$ or $20\%$) and provide them with $60$ more examples from that set. We considered $15$ utterances enough to determine the noise fraction, but used the ratings on $75$ utterances for evaluating realism (see realistic utterance \% in \autoref{tab:noise_data_analysis}).

In \autoref{tab:noise_data_analysis}, we summarize the results of the human evaluation. Column two shows the error injection ratio that was deemed to have more than $95\%$ realistic utterances. We set a high cut-off of $95\%$ to ensure we can make confident statements about the robustness of multilingual models to realistic alterations exhibited in our benchmarks. Hence, Chinese and Japanese (with a realism of $86.2\%$ and $92.3\%$ resp.) are omitted in our benchmarks. The last two columns highlight examples deemed as realistic and unrealistic by human experts with the noised tokens highlighted in orange.

Given the sentence length and similarity in task types, we use the error injection percentage determined to be the most realistic for \ma~as the error injection percentage for \ms~and Wiki-ann. For XNLI, experts deemed higher noise injection ratios (of $>0.05$) to be unrealistic ($15\%$ for 0.1, $27\%$ for 0.2) because (1) the premise, usually much longer than sentences in \ma, had (impractically high) number of noise tokens, and (2) the classification label (implies/neutral/contradicts) sometimes changed with large noise additions. Thus, for \nli, we choose $0.05$ to be the default noise injection ratio. Finally, one expert noted the Turkish data for \ma~lacked many diacritic characters, muddling the distinction between noise injected by our procedure and existing misspellings; hence, it was ignored.

\begin{figure}[t]
    \centering
    \includegraphics[
    width=\columnwidth,
    trim={0.2cm 1.75cm 1.5cm 2cm},
    clip
    ]{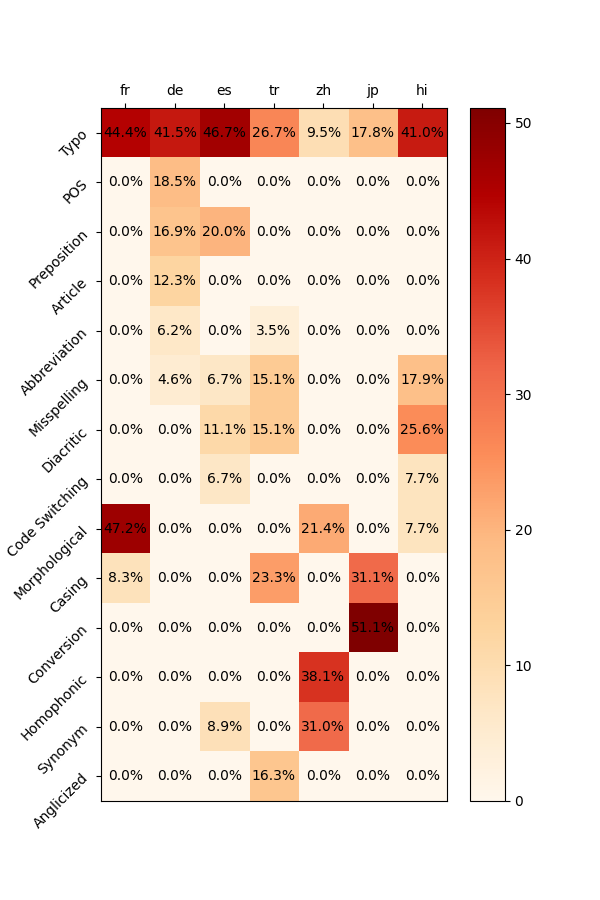}
    \caption{The column-wise color density (which adds up to one) shows the percentage of a different noise types observed for a particular language. The row-wise values show that some noise types (eg. homophononic) is only present for a single language (eg. zh).}
    \label{fig:noise_types}
    \vspace{-10pt}
\end{figure}

In \autoref{fig:noise_types}, we list the noise-types identified by our experts in different languages. While certain noise-types, such as typographical errors, misspellings are common across multiple languages, there are various language-specific noise-types, such as homophonic errors (for zh), Kanji conversion errors (for ja), anglicization (for tr) (we showcase some examples in \autoref{sec:noise-types}). Given disjoint noise types across languages, we expect that augmentation with errors seen in English (using approaches proposed by prior works) will generalize better to languages that share error types.

\section{Robust Contrastive Pre-training (RCP)}
\label{sec:contast}

\begin{table*}[t]
\small
\centering
\begin{tabular}{lcccccc}
\toprule
\textbf{Dataset} & \textbf{Task} & \textbf{Size (training)} & \textbf{Languages} & \textbf{Epochs} & \textbf{Learning Rate} & \textbf{Seeds} \\
\midrule
\ma~\cite{xu-etal-2020-end}             & IC/SL         & 5k                       & de,en,es,fr,hi       & 80                                  & 1E-04                        & 5                                  \\
\quad+ training data aug.  &               & 18k                      & de,en,es,fr,hi       & 20                                  & 1E-04                        & 5                                  \\
\ms            & IC/SL         & 13k                      & en,es,fr,hi          & 40                                  & 1E-04                        & 5                                  \\
\quad+ training data aug.  &               & 72k                      & en,es,fr,hi          & 10                                  & 1E-04                        & 5                                  \\
\wa~\cite{pan-etal-2017-cross}         & NER           & 20k                      & de,en,es,fr,hi,tr    & 3                                   & 2E-05                        & 5                                  \\
\nli~\cite{conneau-etal-2018-xnli}         & NLI           & 392k                     & de,es,fr,hi,tr       & 5                                   & 2E-05                        & 5        \\
\bottomrule
\end{tabular}
\caption{Data-set characteristics and hyper-parameters for our experiments.}
\label{tab:stat_n_hp}
\end{table*}

\begin{figure}
    \centering
    \includegraphics[width=0.49\textwidth]{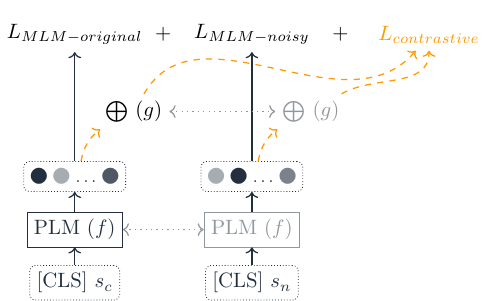}
    \caption{Loss function for fine-tuning a Pretrained Language Model (PLM) using Robust Contrastive Pretraining (RCP).}
    \label{fig:rcp_architecture}
\end{figure}

\paragraph{Motivation and Approach}
While {\em task-time} data augmentation (aka adversarial training) has been effective to boost the robustness of pre-trained models for English, we face two major challenges-- (1) lack of supervised multilingual training data in our zero-shot setting, and (2) lack of approaches to synthetically generate noise data for non-English languages. We overcome these with a multilingual data-augmentation approach at {\em pre-training time} that uses the multilingual Wikipedia edit corpus to expose our models to human errors during pre-training. Here, the need of ex-situ injection of noise (for test-data creation \cref{sec:ds_creation}) is unnecessary as our edit corpus contains pairs of similar sentences, i.e. a version of the sentence before and after revision by a Wikipedia contributor (\cref{sec:wiki_mining}).
To encourage the model to align the representations of these two sentences in the encoder's output space, we use a contrastive loss term (see \autoref{fig:rcp_architecture}). Building on previous work on contrastive learning \cite{giorgi-etal-2021-declutr}, Robust Contrastive Pre-training (RCP) considers the original and edited version of a sentence as positive examples and other unrelated sentences as the negative examples.

Similar to \citet{giorgi-etal-2021-declutr} and \citet{reimers-gurevych-2019-sentence}), we map variable length sentences to fixed-length embeddings with a pooler ${e}_i = g(f(s_i))$, where $f(\cdot)$ is a transformer encoder, and  $g(\cdot)$ is the mean of the token-level embeddings. 
Given a batch of $N$ (noisy, clean) sentence tuples, we set our original sentence $s_{c}$ as the anchor and the noisy version $s_{n}$ as the corresponding positive pair. \footnote{One obvious choice would be for clean sentence with index $2i$, the noisy sentence has index $2i-1$.} Other sentences in the batch (i.e. $\neq s_n$) are deemed to be negative examples.
We consider the InfoNCE/NT-Xent loss \cite{sohn2016improved} for our  per-example contrastive loss:
\begin{align}
    \ell(i, j) &= -\log \frac{\exp(\text{sim}(e_{i}, e_{j}))}{\sum_{i \ne k} \exp(\text{sim}(e_{i}, e_k) / \tau)}
\end{align}%
where $\text{sim}(u, v) = u^Tv / ||u||_2||v||_2$ denotes the cosine similarity of two vectors $u$ and $v$ %
and \(\tau > 0\) denotes the temperature hyper-parameter. Thus, our final contrastive loss function is
\begin{align*}
    L_{\text{contrastive}} &= \sum_{i=1}^{N} \ell(c, n) + \ell(n, c)
\end{align*}%
We additionally use the standard MLM loss at pre-training time, masking 15\% of the input tokens of every sentence (i.e.\ both noisy and clean) independently. Therefore, our final loss function is
\[
L = L_{\text{contrastive}} + L_{\text{MLM-noisy}} + L_{\text{MLM-original}}
\]%
$L_{\text{MLM-original}}$ is the MLM loss on original sentences, and ensures the model does not `forget' its original pre-training task. $L_{\text{MLM-noisy}}$ is the MLM loss on noisy sentences, and can be thought of as data-augmentation at pre-training time.

\begin{table*}[t]
\centering
\small
\begin{tabular}{lcllllll}
\toprule
    Model  & Original/ Noisy & \multicolumn{2}{c}{\ma} & \multicolumn{2}{c}{\ms} & Wiki-ann       & XNLI           \\
      &     & IC\%          & SL-F1          & IC\%         & SL-F1          & NER-F1         & NLI\%        \\
      \midrule
XLM-R$_{\textrm{base}}$  & Original   & \textbf{90.68}  & \textbf{71.45} & \textbf{92.93} & \textbf{68.01} & \textbf{74.14} & \textbf{76.69} \\
      & Noisy   & \textbf{89.65}  & \textbf{62.3}  & \textbf{90.46} & \textbf{61.63} & \textbf{69.48} & \textbf{74.38} \\
mBERT & Original  & 86.29           & 64.95          & 78.65          & 59.05          & 73.92          & 70.82          \\
      & Noisy   & 85.42           & 55.17          & 75.35          & 53.71          & 69.38          & 68.44         \\

      \bottomrule
\end{tabular}
\caption{Performance of pre-trained multilingual models on the four multilingual datasets averaged across languages and 5 seeds. \xlmr{} outperforms mBERT on Original and Noisy test data across all metrics.}
\label{tab:pretrained}
\vspace{-5pt}
\end{table*}

\paragraph{Pre-training Details}
Following the Domain Adaptive Pre-Training (DAPT) approach \cite{gururangan-etal-2020-dont}, we start with an existing multilingual pre-trained model and fine tune it with our RCP objective. Unlike DAPT, we are not interested in specializing in a particular domain, but in increasing robustness to errors. As mentioned before, we use (unfiltered) pairs of correct/incorrect sentences from the multilingual Wikipedia archive and include sentences from the Lang8 corpus.\footnote{ \url{https://sites.google.com/site/naistlang8corpora/}} The  Lang8 corpora consists of a smaller number of sentences compared to the Wikipedia corpus, but proves to be apt for our purpose; it consists of pairs of sentences-- one written by a non-native speaker who is learning the language (eg. ``As the winter is coming, I'm getting to feel better.'') and a re-write of this sentence by a native speaker (eg. ``as the winter is coming, I'm starting to feel better.''). More details about the individual corpora can be found in \autoref{sec:datastats}.

We note that the pre-training corpus is not exactly the same set of sentences used to construct our noise dictionaries in \cref{sec:wiki_mining}. In this case, the only criteria for inclusion is a length difference of $<5$ tokens, and a relative edit-distance of $30\%$ of the shorter sentence (see \cref{sec:datastats} for more details). Hence, we incorporate training data from the corpora that exhibit changes beyond simple typos (such as paraphrasing, sentence-level morphological variance) in the pre-training stage.\footnote{Unfortunately, the benefit of including sentence-level noise in the pre-training phase is not directly examined by our benchmarks, which focus more on word-level noise.}

Similar to \citet{gururangan-etal-2020-dont}, we fine tune for 25k steps with a batch size of 2048 sentences to create two pretrained models-- one with $\mathcal{L}_{\text{contrastive}} + \mathcal{L}_{\text{MLM-noisy}} + \mathcal{L}_{\text{MLM-clean}}$ (referred to as Robust Contrastive Pre-training or RCP) and an ablation without the contrastive term, i.e. $\mathcal{L}_{\text{MLM-noisy}} + \mathcal{L}_{\text{MLM-clean}}$. The latter setting represents a pure (pre-training time) data augmentation approach such as \citet{tan-etal-2020-morphin} (termed {\em p(aug)} in \autoref{tab:noise-compare-methods}). See \autoref{sec:hyperparams} for more hyper-parameters and settings.

\begin{table*}[t]
\small
\centering
\begin{tabular}{lllllllll}
\toprule
Task              & Metric  & XLMR & \multicolumn{1}{p{1cm}}{XLMR +p(aug)} & \multicolumn{1}{p{1.4cm}}{XLMR +t(En-aug)} & \multicolumn{1}{p{1cm}}{XLMR +RCP (Ours)} & \multicolumn{1}{p{1.2cm}}{XLMR +RCP+t (Ours)} & Gain \\
\midrule
\ma~& IC\% & 89.65                    & 93.10                                                    & 91.26                                                        & 93.80                                                                                        & \textbf{94.57}                                                                                      & \cellcolor[HTML]{EBFBEA}+4.92                                            \\
             & SL-F1  & 62.30                     & 67.47                                                   & 74.62                                                        & 67.45                                                                                        & \textbf{80.68}                                                                                      & \cellcolor[HTML]{EBFBEA}+18.38                                           \\
\ms~ & IC\%   & 90.46                    & 93.98                                                   & 91.60                                                         & 93.79                                                                                       & \textbf{94.53}                                                                                      & \cellcolor[HTML]{EBFBEA}+4.07                                            \\
             & SL-F1  & 61.63                    & 66.67                                                   & 66.44                                                        & 67.69                                                                                       & \textbf{70.20}                                                                                       & \cellcolor[HTML]{EBFBEA}+8.57                                            \\
Wiki-ann     & NER-F1 & 69.48                    & 72.32                                                   & -                                                            & \textbf{72.37}                                                                              & -                                                                                                   & \cellcolor[HTML]{EBFBEA}+2.89                                            \\
XNLI         & NLI\%  & 74.38                    & 74.83                                                   & -                                                            & \textbf{75.06}                                                                              & -                                                                                                   & \cellcolor[HTML]{EBFBEA}+0.68                 \\
\bottomrule
\end{tabular}
\caption{Average performance across languages and five seeds. We abbreviate the baselines, multi-lingual pre-training time augmentation as p(aug), and English task-time (aggregate) data augmentation as t(En-aug). `RCP' stands for `Robust Contrastive Pre-training', and `RCP + t' means combining RCP with task-time data augmentation. `Gain' refers to the increase in performance of the best method vs. \xlmr.}
\label{tab:noise-compare-methods}
\vspace{-5pt}
\end{table*}

\section{Experiments and Results}

We divide this section into three parts. In \cref{sec:robustness-models}, we analyze the robustness of popular multilingual language models in the zero-shot cross-lingual setting. In \cref{sec:rob_con}, we show that Robust Contrastive Pre-training (RCP) improves the robustness of existing baselines on noisy test-data for all tasks-- joint intent classification and slot labeling (IC-SL), Slot-Labeling (SL) Named Entity Recognition (NER) and Natural Language Inference (NLI)-- and (not only maintains but) improves performance on the original test data. Finally, in \cref{sec:failure_mode_analysis}, we conduct failure mode analysis for \ma~and discover that the model trained with RCP makes more explicable sequence-labeling errors (for slot-value prediction)  in comparison to existing baselines.

\paragraph{Setup}
We consider four datasets (shown in \autoref{tab:stat_n_hp}) and four metrics for evaluation. Two of these metrics consider sentence classification accuracy-- Intent classification Accuracy (IC\%) for the goal-oriented dialog text datasets \ma~and \ms, and classification accuracy (NLI\%) for XNLI. We also consider F-score for sequence-labeling tasks-- Slot Labelling (SL-F1) for \ma~and Multi-SNIPS++ and Named Entity Recognition (NER-F1) for Wiki-ann. \autoref{tab:stat_n_hp} shows the languages present in the noisy test data and the size of the English training data used in our zero-shot cross-lingual setting. Note that for task-time data augmentation, we follow the strategy of {\em aggregate noise augmentation} proposed in \cite{sengupta-etal-2021-robustness} for English, which involves augmenting training data with a variety of synthetic noise types such as typos, making words ALLCAPS, abbreviations etc. As this augmentation procedure increases the size of the training data-set $\approx 3.5$ times for \ma~and $\approx 5.5$ times for \ms, we find that training for fewer epochs yields the best results.

\subsection{Robustness of Multilingual Models}
\label{sec:robustness-models}

We compare the robustness of two popular pre-trained language models-- \xlmr~and multi-lingual BERT in the zero-shot cross-lingual setting.\footnote{We also considered Canine-c \cite{DBLP:journals/corr/abs-2103-06874}, a token-free baseline, but observed poor performance compared to \xlmr~and BERT on IC-SL tasks (see \autoref{fig:full_pretrained}).} In this setup, we fine-tune the pretrained language models on the task-training data in English and test (zero-shot) on multilingual test sets. The results reported in \autoref{tab:pretrained} are averaged across multiple languages for brevity (and provide a detailed breakdown in \autoref{sec:detailed_results}). A secondary goal of this experiment was to decide which pre-trained model to use for further experiments and we base our judgements on twelve metrics across four datasets.

Noise always leads to a decrease in performance. On average, the accuracy of both models decreases by $\approx 2\%$ for sentence-level tasks (IC\%, NLI\%), and by $\approx 6.6$ F1-points on sequence-labeling tasks (SL, NER), on noisy data compared to clean data. This can perhaps be explained by the ability to ignore a particular token for sentence-level tasks, whereas every token, including noisy ones, need to be assigned a label for sequence-labeling tasks.

We observe that \xlmr~outperforms mBERT on all the twelve metrics. For sentence-level tasks (i.e. IC\%, NLI\%), \xlmr~outperforms mBERT by $8.43\%$ on average on the noisy test-sets and for sequence-tagging tasks (i.e. SL, NER), \xlmr~outperforms mBERT by $5.1$ F1-points. In general, \xlmr~also seems to be a model better suited for these tasks in the zero-shot cross-lingual setting, as we also see similar gains when using \xlmr~on the clean data.

Breaking the results down by language (see \autoref{sec:detailed_results} for detailed results), \xlmr~outperforms mBERT on average across all languages. Specifically \xlmr~outperforms mBERT on German (in 6/8 metrics), on Spanish (10/10), on French (8/12), on Hindi (12/12), and on Turkish (4/4). As German is missing in \ma~and Turkish is only present in WikiANN and XNLI among the four datasets, the overall number of metrics is less than 12 for these two languages. Given these results, we consider \xlmr~as the baseline multilingual language model in the rest of our experiments.

\begin{table}[t]
\small
\centering
\begin{tabular}{lllll}
\toprule
Task             & Metric & XLMR  & Ours & \multicolumn{1}{l}{Gain}      \\
\midrule
\ma~& IC\%   & 90.68 & 95.32                                                           & \cellcolor[HTML]{EBFBEA}+4.64  \\
             & SL-F1  & 71.45 & 84.07                                                           & \cellcolor[HTML]{EBFBEA}+12.62 \\
\ms~       & IC\%   & 92.93 & 95.66                                                           & \cellcolor[HTML]{EBFBEA}+2.73  \\
             & SL-F1  & 68.01 & 74.39                                                           & \cellcolor[HTML]{EBFBEA}+6.38  \\
Wiki-ann     & NER-F1 & 74.14 & 76.34                                                           & \cellcolor[HTML]{EBFBEA}+2.2   \\
XNLI         & NLI\%  & 76.69 & 76.75                                                           & \cellcolor[HTML]{EBFBEA}+0.06 \\
\bottomrule
\end{tabular}
\caption{Comparison of our RCP method with the baseline \xlmr{} model on the original (clean) test data.}
\label{tab:clean-compare-methods}
\vspace{-5pt}
\end{table}

\begin{table*}[t]
\centering
\small
\begin{tabular}{p{2cm}p{7.5cm}p{3.1cm}}
\toprule
Error Type & Utterance & Slot-lables \\
\midrule
Hallucination & 
{\em Ichs} brauche einen Flug von Memphis nach Tacoma, der {\em \textbf{uber}} Los Angeles fliegt
&
\sai{\faCheck} O (über) \newline
\sm{\faExclamation} airline\_code
\\
Contextual &
Zeige {\em mit der} Erste-Klasse und Coach-Flüge vom \textbf{JFK} nach Miami
&
\sai{\faCheck} fromloc.airport\_code \sm{\faExclamation} toloc.airport\_code
\\
\bottomrule
\end{tabular}
\caption{Examples of slot labeling errors in German-- errors are in {\em italics}; misclassified tokens are \textbf{bold}.}
\label{tab:de-eg}
\end{table*}

\subsection{Robust Contrastive Pre-training Results}
\label{sec:rob_con}

To showcase the efficacy of our RCP approach, we compare our approach to a popular multilingual model \xlmr, which performed best in the previous section, and two augmentation solutions that were proposed earlier and shown to improve robustness of English language models to real-world noise. 
First, we consider a pre-training time data augmentation approach, similar to \citet{tan-etal-2020-morphin}, by continuing to pre-train \xlmr{} on \textit{noisy} multilingual data; see section~\ref{sec:contast}.
Next, we consider augmenting task-time data with a combination of various noise types, following \citet{sengupta-etal-2021-robustness} that shows using this aggregate data augmentation during task-time finetuning improved performance on both noisy and clean data for IC-SL tasks like ATIS and SNIPS. For the latter, we treat it as a baseline for zero-shot cross-lingual transfer for the dialog-datasets--\ma~and \ms-- and also combine it with our pre-training time approaches.

As shown in \autoref{tab:noise-compare-methods}, our approach can improve the performance of current multilingual models across all 4 tasks and datasets. For the multilingual goal-oriented dialog datasets, our approach coupled with task-time augmentation outperforms all the other methods. We observe that the gain for SL tasks is higher than that obtained for IC tasks. Although we analyze the SL results further in \cref{sec:failure_mode_analysis}, we highlight that IC accuracy is less affected by noise than SL F1; this provides more headroom for improving SL metrics. The highest gains are observed for Hindi where the \xlmr~model has the worst SL performance on noisy data (42.86 for \ma, 36.93 for \ms).
Likewise, we also observe improvement on XNLI\% and NER-F1; the largest improvement is again seen on the noisy data for Hindi. Overall, the gain on sequence-labelling tasks is larger than the gain on sentence-level classification tasks.

{\em Does this improvement on noisy data come at the cost of worse performance on clean data?} In \autoref{tab:clean-compare-methods}, we show that the best performing models shown in \autoref{tab:noise-compare-methods} (XLMR+RCP+t for \ma~and \ms, and XLMR+RCT for \wa~and \nli) also improve the performance on clean test data. Further, the magnitude of growth seen on clean data is like the ones seen on the noisy test data. For slot-labeling errors, we observe a particular kind error which occurs on both clean and noisy data that our model mitigates; we provide more details on this in the next section. For IC and XNLI, we found no specific error pattern that distinguishes between \xlmr{} and our model. Thus, we believe that our approach mostly improves the overall quality of the model’s representation rather than just its downstream robustness. In the future, one can consider if an upper bound on model quality exists beyond which the tension between accuracy on clean data and robustness to real-world noise emerges \cite{tsipras2018robustness}.

Finally, we note that beyond improving performance on clean and noisy data, our approach reduces the \textit{disparity} in performance between the clean and noisy test sets. For \ma, the disparity reduces by 0.3\% for IC\% and 5.76 for SL-F1; for \ms, it reduces by 1.34\% for IC\% and 2.19 for SL-F1; for \wa, it reduces by 0.68 for NER-F1; and for \nli, it reduces by 0.9\% for NLI\%.

\begin{figure}[t]
    \centering
    \includegraphics[width=0.98\columnwidth]{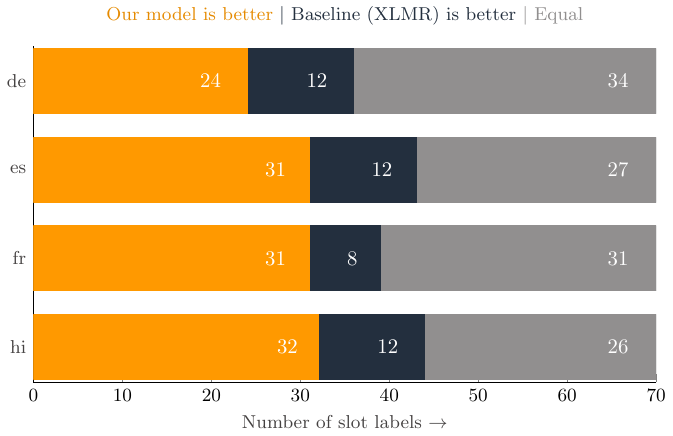}
    \caption{Comparing the number of slot labels for which our model {\em vs.} the baseline performs better.}
    \label{fig:sl_analysis}
\end{figure}

\subsection{Result Analysis}
\label{sec:failure_mode_analysis}

Given the large improvement seen on sequence labeling tasks, we zoom in on the SL metrics for \ma. In \autoref{fig:sl_analysis}, we show the number of slot labels on which our method outperforms (has fewer misclassifications than) the baseline, vice versa, and where they perform equally. Our method clearly out-performs the baseline on at least twice the number of slot-labels-- $2\times$ better on German, $\approx 2.6\times$ times on Spanish and on Hindi, and $\approx 4\times$ on French. Across all languages, our model always outperforms \xlmr~on eight slot-labels. These slots correspond to relative times (`leaves in the \textit{evening}’), relative dates (`traveling the \textit{day after tomorrow}’), relative costs (`\textit{cheap} flights’), meal names (`flights that offer \textit{breakfast}’), and carrier tokens/non-slot values (`\textit{that offer} breakfast’). We postulate these slot values are more common in the pre-training data compared to proper nouns such as airline, airport or city names and thus, understood in noisy contexts. In turn, variations of these words are mapped closer in the embedding space and the classifier is more robust to such errors.

Upon further analysis, we observe two distinct patterns-- (1) reduction in \textit{hallucination errors}, i.e. errors where an irrelevant carrier phrase token is labeled to be a slot value, and (2) errors become more contextual-- misclassification is to related classes (see examples in \autoref{tab:de-eg}).

\begin{table}[t]
\small
\centering
\begin{tabular}{lllllll}
\toprule
N/O   & Model                                                                           & de  & es  & fr  & hi  \\
\midrule
Noisy & XLMR                                                                            & 315 & 358 & 413 & 671 \\
 & XLMR+RCP+t & 21  & 123 & 33  & 204 \\
Original & XLMR                                                                            & 208 & 262 & 334 & 460 \\
 & XLMR+RCP+t & 19  & 106 & 22  & 180 \\
\bottomrule
\end{tabular}
\caption{Reduction in hallucination error (i.e. model identifies irrelevant tokens as a slot value) counts.
}
\label{tab:hallucination}
\end{table}

\begin{table}[t]
\small
\centering
\begin{tabular}{p{3.5cm}rrrr}
\toprule
Languages       & de & es & fr & hi \\
\midrule
\textbf{(r1)} Top-confusion changes to no-label (w/ RCP) & 7  & 8  & 6  & 17 \\
\textbf{(r2)} Confusions becomes more explicable (w/ RCP) & 8  & 3  & 3  & 4 \\ %
\bottomrule
\end{tabular}
\caption{Number of slot-labels that our model misclassified to (r1) a no-slot or (r2) a more explicable slot-label.}
\label{tab:sl-top-mis}
\vspace{-5pt}
\end{table}

In \autoref{tab:hallucination}, we highlight the distribution of hallucination errors and observe that the number of carrier phrase tokens that the baseline \xlmr~misclassifies as a slot-value reduces (by $>10\times$ for German and French, and $\approx$2-3$\times$ for Hindi and Spanish) with our approach on both the original and the noisy test data. This observation aligns with our initial reasoning that the contrastive loss term at pre-training time helps the model develop a better understanding of non-slot words as the model learns to identify such words (and their noisy forms) in both linguistically correct and noisy contexts. Note that the latter signal is missing for the \xlmr~baseline.

For a subset of the slot labels, the class to which it was misclassified (with the highest frequency) differed between the \xlmr~baseline and our model. In \autoref{tab:sl-top-mis}, we highlight two scenarios where the most-confused label changed from (r1) an incorrect slot label (eg. {\em meal\_code} $\rightarrow$ {\em airline\_code}) to no-label (i.e. {\em meal\_code} $\rightarrow$ O), and (r2) from an inexplicable slot label ({\em state\_code} $\rightarrow$ {\em transport\_type}) to a more explicable one ({\em state\_code} $\rightarrow$ {\em state\_name}) when the RCP method is used (we use the explicable/inexplicable terminology of \citet{olmo2020not}). Thus, our approach inadvertently improves the explicability of the failures made during slot-labeling. 

\section{Conclusion}

In this paper, we investigate the robustness of pretrained multilingual models in the zero-shot cross-lingual setting on four tasks-- intent classification, slot labeling, named entity recognition, and natural language inference. Given the dearth of existing datasets to benchmark the robustness of existing multilingual models, we develop noisy test data by injecting errors mined from an edit corpus (and conduct expert evaluation for quality assurance). Our identification of noise types across various languages motivates the necessity of language specific investigation in the future. Finally, demonstrate existing baselines perform poorly in the presence of noise in the test data and propose Robust Contrastive Pretraining to boost the robustness of these multilingual models.

\section{Ethical Considerations}
\label{sec:ethics}

For the human annotation tasks of (1) identifying language-specific noise types, and (2) ranking their realism, we leveraged the effort of full-time employees at Amazon. The annotators had advanced degrees in linguistics or natural language processing, and were fluent/native in the languages they annotated. Amazon compensated them under a competitive industry rate, which is above the minimum hourly pay rate, for their particular job role (which included Applied/Research Scientists, Software/Language Engineers, Linguists, and Language Consultants).

\paragraph{Acknowledgements}
A special thanks to Saab, Batool Haider and M. Saiful Bari for sharing with us the MultiSNIPS dataset. In addition, we want to express our gratitude to members of the AWS AI Lab for their valuable comments, suggestions, and participation in our pilot and human labeling studies (in no particular order)-- Sebastien Jean, Volha Belash, Arshit Gupta, Berk Sarioz, Maansi Shandilya, Raphael Shu, Abhilash Panigrahi, Lorenzo Lambertino, and Yi Zhang. Finally, we are grateful to the anonymous reviewers who have helped us improve this paper. 

\section{Limitations}

\subsection{The Umbrella of Realistic Noise}
`Realistic noise' is too abstract a category. We mostly concern ourselves with real-world errors and their corrections appearing in existing corpora (with criteria like a small character-level edit distance). But this could include things like better paraphrasing, use of more appropriate synonyms or morphology that can be viewed as language variation rather than noise; this could be one reason we notice improvements on the original (i.e. un-noised) test data. Yet, to distinguish ourselves from the terminology of synthetic or adversarial noise, we choose this (imperfect) terminology of real-world/realistic noise as in \citet{sengupta-etal-2021-robustness} to bracket all our noise types under a single class.

\subsection{Language Choice and Diversity}
This work considers (relatively) high-resource languages. This makes it easier for us to find publicly available corpora from where we can mine error/correction data and use it to improve the model's understanding of errors and, in turn, boost their robustness to real-world noise. But this is only the first step towards developing an understanding of noise phenomena in languages beyond English, bench-marking multi-lingual model performance in such settings, and improving their robustness. Further, we do notice that Hindi (and, to some extent, Turkish) are relatively low resource languages when it comes to pre-training data (see \autoref{tab:datastats} in Appendix). We hope future work builds on this and explores a greater variety of languages.%

\subsection{Zooming-in on Individual Tasks}

Many of our human studies are based on a subset of datasets (eg. MultiATIS, XNLI). It is possible individual tasks and further, individual datasets need more fine-grained human attention. Given language expertise for several datasets and several languages is difficult/costly, we made the choice to concentrate on a smaller number of datasets in order to provide a more rigorous analysis. We hope future work can expand the number of tasks and datasets covered so we have a more comprehensive analysis of how multilingual noise affects pre-trained models. %

\bibliography{anthology,custom}
\bibliographystyle{acl_natbib}

\appendix

\onecolumn

\section{Examples of Noise/Errors in the test set}
\label{sec:noise-types}
\begin{figure}[h]
    \centering
    \includegraphics[
    width=0.5\columnwidth,
    trim={0cm 1.75cm 1.5cm 2cm},
    clip
    ]{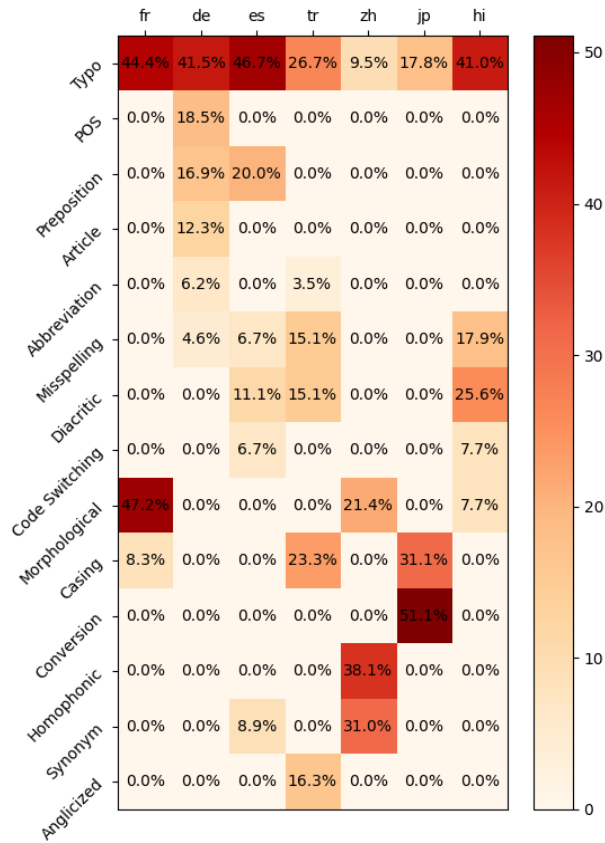}
    \caption{Noise types seen across various languages.}
    \label{fig:noise_types_appendix}
    \vspace{-10pt}
\end{figure}

In this section, we highlight an example of some of the unique noise types observed for certain languages shown in \autoref{fig:noise_types_appendix}.

\subsection{Typographic Errors (Typos)}

Two examples follow for Hindi and Chinese, where experts evaluated based on the Indic Script and the Pinyin keyboards (which is what they use regularly) respectively.

\includegraphics{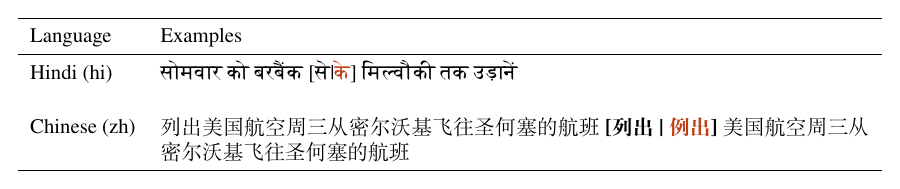}

\subsection{Preposition Errors}

We noticed language experts tagged preposition errors for French and German. Examples follow:

\includegraphics{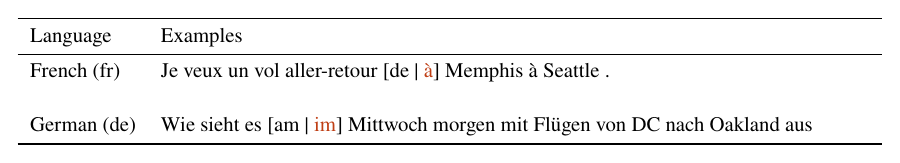}

\subsection{Diacritic Errors}

Some languages use diacritic characters; although even these diacritics may greatly differ depending on script. Examples from Hindi and Spanish follow.

\includegraphics[]{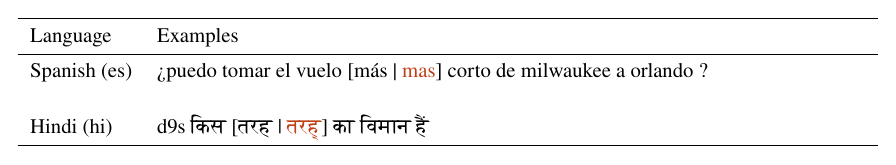}

\subsection{Conversion Errors}

Kanji conversion error. This error was unique to the Japanese language. Examples follow.

\includegraphics[]{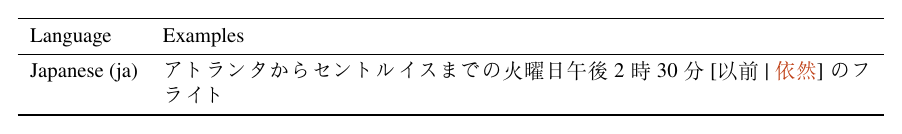}

\subsection{Homophonic Errors}

This error was unique to Chinese. Words with the same pronunciation (potentially with different tones), but different spelling. Examples follow.

\includegraphics[]{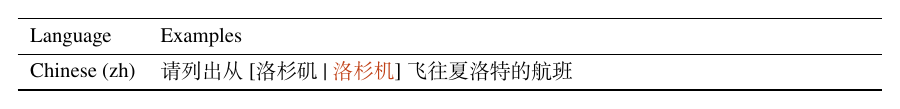}

\subsection{Synonym}

Experts marked these as use of a different synonym in Spanish and Chinese only. Note that such variations may not be erroneous but is still considered a noise given they are not used in the original training/testing data in the given context as much. Examples follow.

\includegraphics[]{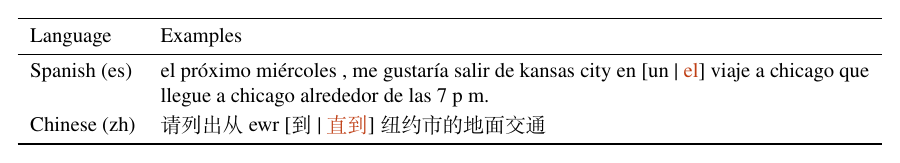}

\subsection{Anglicized}

We observed this errors only for Turkish and noticed that experts marked scenarios where an alphabet in the native script was replaced with a particular one in the latin script. Examples follow (note that Turkish examples are drawn from the \nli~dataset, while the others were drawn from \ma).

\includegraphics[]{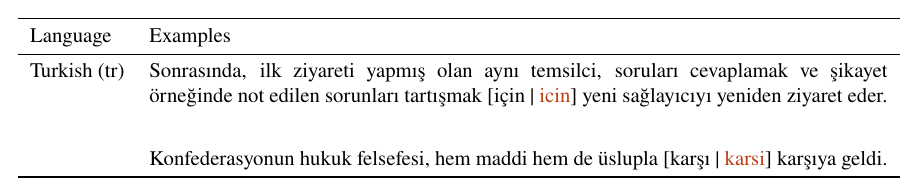}

\clearpage

\section{Chinese and Japanese Edit Mining}
\label{sec:cjk}

Our two character edit distance criteria for obtaining word-level correct-noisy pairs of words does not work well for Chinese characters, including Kanji for Japanese. This is because words are made up of only a small number of characters (relative to e.g. latin scripts). So we can completely change the semantics with only a small character-level edit distance. We therefore used different noise types: Homophonic and Synonym errors for Chinese and Kanji Conversion errors for Japanese, with brief descriptions and examples in \autoref{sec:noise-types}. In order to collect homophonic errors we converted words to pinyin\footnote{Using the \texttt{pinyin} Python package \url{https://pypi.org/project/pinyin/}} (without tone markers) and checked if they were the same in pinyin but different in Chinese characters. To collect synonym noise we labelled words with part-of-speech (POS) tags \footnote{With the \texttt{jieba} Python package \url{https://pypi.org/project/jieba/}.}, and kept words that weren't labeled as nouns, verbs, adverbs, keeping e.g.\ prepositions and conjunctions, with the hope that these would be less likely to involve the kind of big semantic changes you might get with changes to e.g.\ proper nouns like place names.

However this process was largely driven by trial and error and more work is needed to create a principled pipeline that creates a realistic noise dictionary for these languages.%

Finally for Kanji we re-use the criteria of \citet{tanaka-etal-2020-building} as we re-use their dataset of sentence pairs: checking if the two sentences (containing Kanji) have the same reading.

\section{Data Details}
\label{sec:datastats}

\begin{wraptable}{r}{0.5\textwidth}
\centering
\vspace{-12pt}
\small
\begin{tabular}{cccc}
\toprule
Language & Lang8 & Wikipedia & Total \\
\midrule
en & 2.5 & 3.8 & 6.3 \\
de & 0.2 & 13 & 13.2 \\
es & 0.2 & 7.6 & 7.8 \\
fr & 0.2 & 10.7 & 10.9 \\
hi & 0.001 & 0.1 & 0.101 \\
ja & 4.2 & 1 & 5.2 \\
tr & 0.02 & 0.4 & 0.42 \\
zh & 0.6 & 1.9 & 2.5 \\
\bottomrule
\end{tabular}
\caption{Number of sentences (in millions) used for pre-training.}
\label{tab:datastats}
\end{wraptable}

\autoref{tab:datastats} shows the number of Wikipedia and Lang8 sentences (in Millions) we used for fine-tuning the multilingual models in the pre-training stage (\cref{sec:contast}). As stated earlier, the proportion of data obtained from the Lang8 corpus is less than Wikipedia for most languages except English (where it is comparable) and Japanese (where Lang8 has $\approx 4x$ the data compared to the Wikipedia corpus). In general, Hindi (and Turkish) stand out as a relatively low-resource language in our investigation with less than 0.5 Million sentences.

\begin{wraptable}{l}{0.4\textwidth}
\centering
\small
\begin{tabular}{cc}
\toprule
Language & \# Pairs (in Millions) \\
\midrule
en & 0.13 \\
de & 0.33 \\
es & 0.21 \\
fr & 0.27 \\
hi & 0.04 \\
ja & 0.05 \\
tr & 0.25 \\
zh & 0.01 \\
\bottomrule
\end{tabular}
    \caption{Number of Error pairs by language.}
    \label{tab:noisetats}
\end{wraptable}

\autoref{tab:noisetats} lists the number of correct/incorrect pairs (in Millions) used for noise dictionaries to create the test-sets for the various languages (\cref{sec:ds_creation}). Here too, we can observe that the number of corrections are relatively less for Hindi. Interestingly, the number of errors for Chinese are the least although it representation is significantly more compared to Hindi. This low number of errors is inline with our human studies where even the $5\%$ error injection was deemed to be unrealistic; futher, such low pairs of errors also reduced the diversity of our test set, which would eventually results in a lower-quality test-set. Hence, we drop it from our evaluation benchmarks.

\section{Pre-training Settings}
\label{sec:hyperparams}

For our experiments with Robust Contrasting Pretraining (\cref{sec:contast}) and variants we use the following hyperparameters and setup. We train on 4 Nvidea V100 GPUs, with a per-gpu batch size of 8 sentences with a maximum sequence length of 128 tokens, and 64 gradient accumulation steps, for an overall batch size of $64\times8\times4=2048$ sentences. We use a masked language modeling mask probability of $15\%$ and a learning rate of 1e-4 with the Adam optimizer \cite{adam}, and used 16-bit floating point operations. See below for the arguments of the Huggingface transformers \cite{wolf-etal-2020-transformers} masked language modelling script which we modified\footnote{\url{https://github.com/huggingface/transformers/blob/main/examples/pytorch/language-modeling/run_mlm.py}}

\includegraphics[]{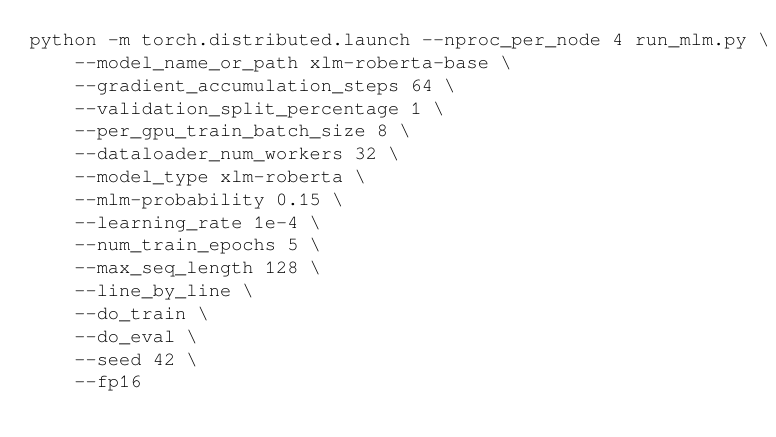}

\section{Per-language Results}
\label{sec:detailed_results}

\autoref{fig:full_pretrained} shows the performance of multilingual models like m-BERT and \xlmr~on  individual languages. We note that the reduction in performance for high-resource language (e.g. German, French, English) is higher than low-resource languages for several settings. To explain this seemingly surprising result, first notice that the metrics on low-resource languages are already bad, even on clean data. Second, the variety of noise seen for low resource languages is less (see \autoref{tab:noisetats}) compared to high-resource settings. Hence, the effect of less diverse noise in low-resource languages doesn't have as large an adverse effect on already poorly performing models.\footnote{We are told a saying goes (coincidentally) in Hindi, {\em mare hue ko kya maroge, saheb?}. It implies you cannot do much (by adding noise) to kill the (model that is already) dead.} 

Another hypothesis, pending future investigation, is that multi-lingual models trained on more high-resource language data overfit to clean test-sets for these languages and fail to generalize better when faced with noise. For low resource languages, the performance on clean data is already poor because of a lack of sufficient language understanding that prevents over-fitting.

\begin{table*}[p!]
\small
\begin{tabular}{lllllllllll}
\toprule
Dataset & Model & Metric & C/N & de     & en     & es     & fr        & hi     & tr     & Avg. \\
\midrule
\textbf{MultiATIS++}  & \textit{XLMR}              & IC\%   & C & 92.4 & 98.7 & 92.0 & 90.6    & 79.6 & -      & 90.7 \\
                      &                           &        & N & 90.9 & 97.6 & 91.8 & 89.5    & 78.4 & -      & 89.6 \\
                      &                           & SL-F1  & C & 74.4 & 96.0 & 73.6 & 70.4    & 42.9 & -      & 71.5 \\
                      &                           &        & N & 67.3 & 82.2 & 68.2 & 65.6    & 38.2 & -      & 62.3 \\
                      &                           &        &   &        &        &        &           &        &        &        \\
                      & \textit{mBERT}           &        &   &        &        &        &           &        &        &        \\
                      &                           & IC\%   & C & 83.3 & 98.3 & 84.7 & 88.8    & 76.3 & -      & 86.3 \\
                      &                           &        & N & 81.2 & 97.6 & 84.3 & 87.9    & 76.1 & -      & 85.4 \\
                      &                           & SL-F1  & C & 59.9 & 96.0 & 65.1 & 69.8    & 33.9 & -      & 65.0 \\
                      &                           &        & N & 51.6 & 78.5 & 60.2 & 64.3    & 31.3 & -      & 55.2 \\
                      & \textit{(XLMR vs mBERT)} &        &   & 4,0    & 1,3    & 4,0    & 4,0       & 4,0    &        &        \\
                      & \textit{}                 &        &   &        &        &        &           &        &        &        \\[0.5em]
 & \textit{Canine-c} & IC\%  & C & 66.32 & 96.51 & 78.41 & 76.06 & 71.55 & - & 77.77 \\
 & \textit{} &       & N & 65.13 & 95.90 & 78.08 & 75.06 & 71.15 & - & 77.06 \\
 & \textit{} & SL-F1 & C & 31.56 & 92.19 & 19.52  & 23.67 & 22.81 & - & 37.95 \\
 & \textit{} &       & N & 32.42 & 78.51 & 20.25 & 24.41 & 22.45 & - & 35.61\\[0.5em]
\textbf{MultiSNIPS++} & \textit{XLMR}  & IC\%   & C & -      & 98.8 & 94.0 & 91.3    & 87.6 & -      & 92.9 \\
                      & \textit{}                 &        & N & -      & 98.4 & 92.4 & 87.0    & 84.1 & -      & 90.5 \\
                      & \textit{}                 & SL-F1  & C & -      & 96.9 & 72.0 & 66.2    & 36.9 & -      & 68.0 \\
                      & \textit{}                 &        & N & -      & 92.7 & 63.3 & 57.7    & 32.8 & -      & 61.6 \\
                      & \textit{}                 &        &   &        &        &        &           &        &        &        \\
                      & \textit{mBERT}           &        &   &        &        &        &           &        &        &        \\
                      & \textit{}                 & IC\%   & C & -      & 98.9 & 88.0 & 88.5    & 39.3 & -      & 78.6 \\
                      & \textit{}                 &        & N & -      & 98.2 & 84.1 & 82.9    & 36.2 & -      & 75.4 \\
                      & \textit{}                 & SL-F1  & C & -      & 96.5 & 65.4 & 59.9    & 14.5 & -      & 59.1 \\
                      & \textit{}                 &        & N & -      & 91.3 & 58.1 & 52.4    & 13.0 & -      & 53.7 \\
                      & \textit{(XLMR vs mBERT)} &        &   &        & 3,1    & 4,0    & 2,2       & 4,0    &        &        \\
                      & \textit{}                 &        &   &        &        &        &           &        &        &        \\[0.5em]
& \textit{Canine-c}  & IC\%  & C & - & 69.39 & 32.88 & 36.39 & 23.28 & - & 40.48 \\
& \textit{} &       & N & - & 69.30 & 32.57 & 34.99 & 23.68 & - & 40.13 \\
& \textit{} & SL-F1 & C & - & 0.89.31 & 24.09 & 23.06 & 6.93 & - & 35.85 \\
&          &       & N & - & 87.86 & 22.3   & 21.49 & 7.02  & - & 34.67
\\[0.5em]
\textbf{WikiANN}      & \textit{XLMR}  & NER-F1 & C & 74.9 & -      & 75.2 & 77.2    & 67.5 & 75.9 & 74.1 \\
                      & \textit{}                 &        & N & 71.6 & -      & 70.0 & 71.1 & 65.1 & 69.5 & 69.1 \\
                      & \textit{}                 &        &   &        &        &        &           &        &        &        \\
                      & \textit{mBERT}           &        &   &        &        &        &           &        &        &        \\
                      & \textit{}                 & NER-F1 & C & 78.6 & -      & 72.1 & 79.5    & 66.2 & 73.1 & 73.9 \\
                      & \textit{}                 &        & N & 75.4 & -      & 67.1 & 74.2    & 63.0 & 67.3 & 69.4 \\
                      & \textit{(XLMR vs mBERT)} &        &   & 0,2    &        & 2,0    & 0,2       & 2,0    & 2,0    &        \\
                      & \textit{}                 &        &   &        &        &        &           &        &        &        \\
\textbf{XNLI}         & \textit{XLMR}    & NLI\%  & C & 76.4 & 84.6 & 78.8 & 77.9    & 69.7 & 72.9 & 76.7 \\
\textbf{}             & \textit{}                 &        & N & 72.6 & 80.7 & 76.4 & 75.7    & 70.3 & 70.6 & 74.4 \\
\textbf{}             & \textit{mBERT}            &        &   &        &        &        &           &        &        &        \\
\textbf{}             & \textit{}                 & NLI\%  & C & 71.1 & 82.0 & 74.9 & 74.2    & 60.5 & 62.2 & 70.8 \\
                      &                           &        & N & 67.5 & 77.9 & 73.1 & 71.8    & 61.5 & 59.1 & 68.4 \\
                      & \textit{(XLMR vs mBERT)} &        &   & 2,0    & 2,0    & 2,0    & 2,0       & 2,0    & 2,0    & \\

\bottomrule
\end{tabular}
\caption{Per-language results of cross-lingual transfer from English data (average of 5 random seeds) across 4 datasets analyzed in \cref{sec:robustness-models} to compare between existing pre-trained multilingual models.}
\label{fig:full_pretrained}
\end{table*}

\end{document}